\documentclass{article} 
\pdfoutput=1
\usepackage{iclr2017_workshop,times}
\usepackage{hyperref}
\usepackage{url}
\usepackage{amsfonts}
\usepackage{amsmath}
\usepackage{bm}
\usepackage{graphicx}
\usepackage{subcaption}
\usepackage[page]{appendix}
\graphicspath{ {./images/} }

\def\Transform{{T}}

\title{Adversarial examples in the physical world}

\author{Alexey Kurakin \\
Google Brain \\
\texttt{kurakin@google.com}
\And
Ian J. Goodfellow \\
OpenAI \\
\texttt{ian@openai.com}
\And
Samy Bengio \\
Google Brain \\
\texttt{bengio@google.com}
}

\DeclareMathOperator*{\argmin}{arg\,min}

\DeclareMathOperator{\sign}{sign}

\begin{document}

\maketitle

\begin{abstract}
Most existing machine learning classifiers are highly vulnerable to adversarial examples.
An adversarial example is a sample of input data which has been modified
very slightly in a way that is intended to cause a machine learning classifier
to misclassify it.
In many cases, these modifications can be so subtle that a human observer does
not even notice the modification at all, yet the classifier still makes a mistake.
Adversarial examples pose security concerns
because they could be used to perform an attack on machine learning systems, even if the adversary has no
access to the underlying model.
Up to now, all previous work has assumed a threat model in which the adversary can
feed data directly into the machine learning classifier.
This is not always the case for systems operating in the physical world,
for example those which are using signals from cameras and other sensors as input.
This paper shows that even in such physical world scenarios, machine learning systems are vulnerable
to adversarial examples.
We demonstrate this by feeding adversarial images obtained from a cell-phone camera
to an ImageNet Inception classifier and measuring the classification accuracy of the system.
We find that a large fraction of adversarial examples are classified incorrectly
even when perceived through the camera.
\end{abstract}

\section{Introduction}

\begin{figure}[h]
  \centering
  \begin{subfigure}[b]{0.24\textwidth}
    \raisebox{0.39\textwidth}{\includegraphics[width=\textwidth]{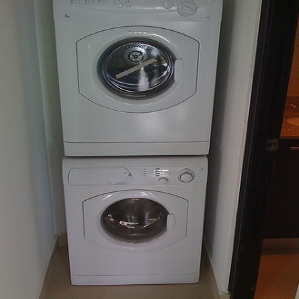}}
    \caption{Image from dataset}
  \end{subfigure}
  \begin{subfigure}[b]{0.24\textwidth}
    \includegraphics[width=\textwidth]{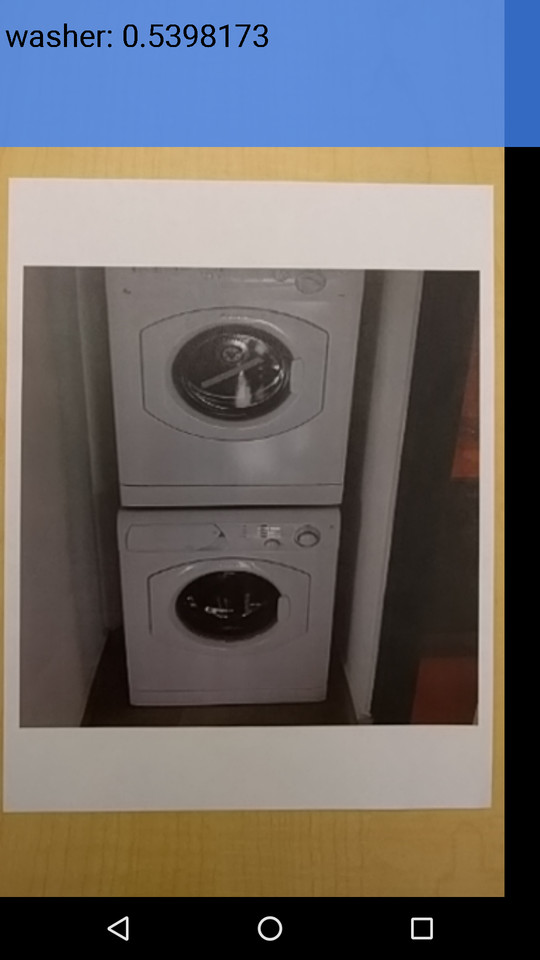}
    \caption{Clean image}
  \end{subfigure}
  \begin{subfigure}[b]{0.24\textwidth}
    \includegraphics[width=\textwidth]{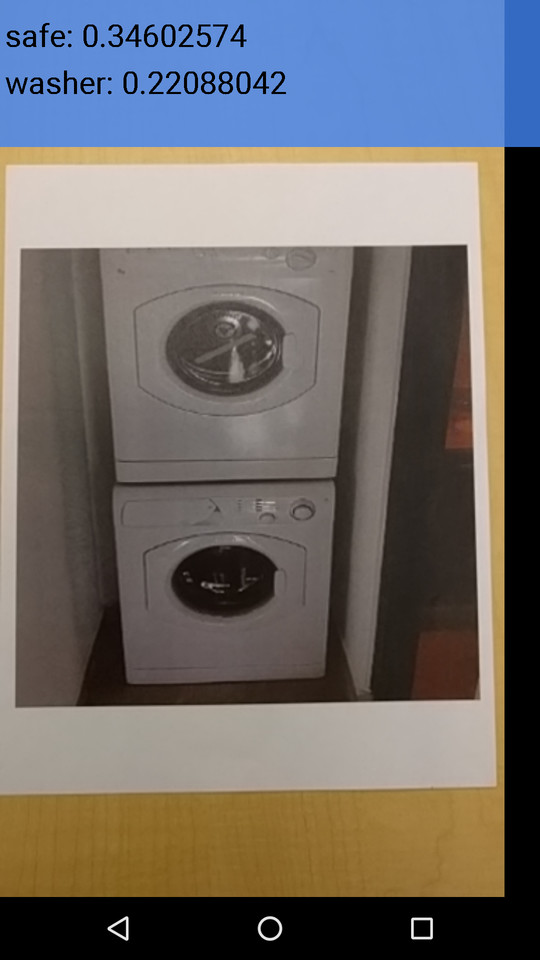}
    \caption{Adv. image, $\epsilon = 4$}
  \end{subfigure}
  \begin{subfigure}[b]{0.24\textwidth}
    \includegraphics[width=\textwidth]{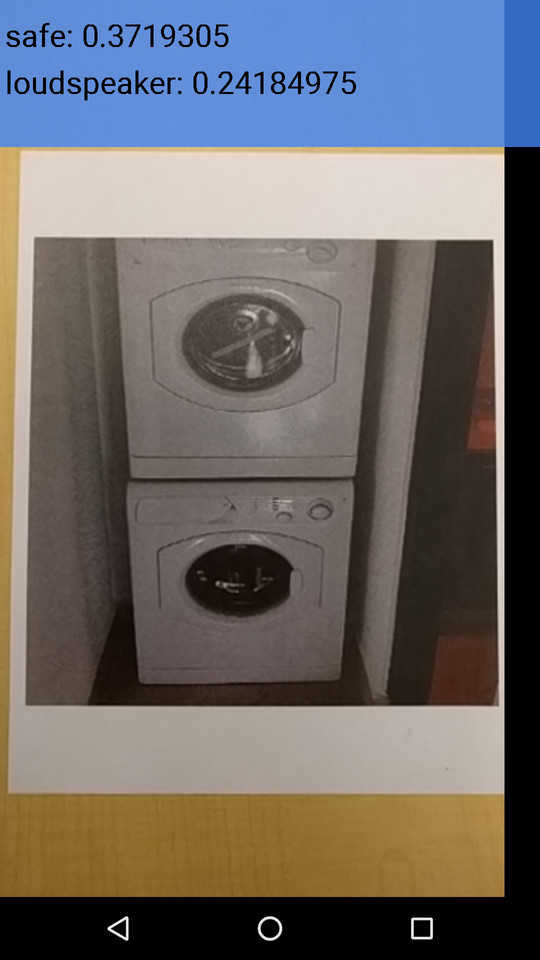}
    \caption{Adv. image, $\epsilon = 8$}
  \end{subfigure}
  \caption{Demonstration of a black box attack (in which the attack is constructed without
  access to the model) on a phone app for image classification using physical adversarial examples.
  We took a clean image from the dataset~(a) and used it to generate adversarial images
  with various sizes of adversarial perturbation~$\epsilon$.
  Then we printed clean and adversarial images and used the TensorFlow Camera Demo app to classify them.
  A clean image (b) is recognized correctly as a ``washer'' when perceived through the camera,
  while adversarial images~(c)~and~(d)~are misclassified.
  See video of full demo at~\url{https://youtu.be/zQ_uMenoBCk}.
  }\label{fig-phone-app-demo}
\end{figure}

Recent advances in machine learning and deep neural networks
enabled researchers to solve multiple important practical problems
like image, video, text classification and others~\citep{Krizhevsky-2012,Hinton-Speech-2012,Bahdanau-et-al-ICLR2015-small}.

However, machine learning models are often vulnerable to 
adversarial manipulation of their input intended to cause
incorrect classification \citep{dalvi2004adversarial}.
In particular, neural networks and many other categories of machine
learning models are highly vulnerable to
attacks based on small modifications of the input to the model
at test time
\citep{biggio2013evasion,Szegedy-ICLR2014,Goodfellow-2015-adversarial,Papernot-2016-transferability}.

The problem can be summarized as follows.
Let's say there is a machine learning system $M$ and input sample $C$ which we call a clean example.
Let's assume that sample $C$ is correctly classified by the machine learning system, i.e. $M(C) = y_{true}$.
It's possible to construct an adversarial example $A$ which is perceptually indistinguishable 
from $C$ but is classified incorrectly, i.e. $M(A) \ne y_{true}$.
These adversarial examples are misclassified far more often than examples that have been perturbed
by noise, even if the magnitude of the noise is much larger than the magnitude of the adversarial
perturbation \citep{Szegedy-ICLR2014}.

Adversarial examples pose potential security threats for practical machine learning applications.
In particular, \citet{Szegedy-ICLR2014} showed that an adversarial example that was designed
to be misclassified by a model $M_1$ is often also misclassified by a model $M_2$.
This adversarial example transferability property means 
that it is possible to generate adversarial examples
and perform a misclassification attack on a machine learning system without access
to the underlying model.
\citet{Papernot-MGJCS16} and \citet{Papernot-2016-transferability} demonstrated such
attacks in realistic scenarios.

However all prior work on adversarial examples for neural networks
made use of a threat model in which the attacker can supply
input directly to the machine learning model.
Prior to this work, it was not known whether adversarial examples would remain
misclassified if the examples were constructed in the physical world and observed
through a camera.

Such a threat model can describe some scenarios in which attacks can take
place entirely within a computer, such as as evading spam filters or
malware detectors \citep{biggio2013evasion,nelson2008exploiting}.
However, many 
practical machine learning systems operate in the physical world.
Possible examples include but are not limited to: robots perceiving world through cameras and other sensors,
video surveillance systems, and mobile applications for image or sound classification.
In such scenarios the adversary cannot rely on the ability of fine-grained per-pixel modifications of the input data.
The following question thus arises:
is it still possible to craft adversarial examples and perform adversarial attacks on machine learning systems
which are operating in the physical world and perceiving data through various sensors, rather than digital representation?

Some prior work has addressed the problem of physical attacks against machine
learning systems, but not in the context of fooling neural networks by making
very small perturbations of the input.
For example, \citet{carlini2016} demonstrate an attack that can create audio
inputs that mobile phones recognize as containing intelligible voice commands,
but that humans hear as an unintelligible voice.
Face recognition systems based on photos are vulnerable to replay attacks,
in which a previously captured image of an authorized user's face is presented to the camera
instead of an actual face \citep{smith2015face}.
Adversarial examples could in principle be applied in either of these physical domains.
An adversarial example for the voice command domain would consist of a recording
that seems to be innocuous to a human observer (such as a song)
but contains voice commands recognized by a machine learning algorithm.
An adversarial example for the face recognition domain might consist of
very subtle markings applied to a person's face, so that a human observer
would recognize their identity correctly, but a machine learning system
would recognize them as being a different person.
The most similar work to this paper is \citet{Sharif16AdvML},
which appeared publicly after our work but had been submitted to a conference
earlier. \citet{Sharif16AdvML} also print images of adversarial examples on
paper and demonstrated that the printed images fool image recognition systems
when photographed.
The main differences between their work and ours are that: (1) we use a cheap
closed-form attack for most of our experiments, while \citet{Sharif16AdvML}
use a more expensive attack based on an optimization algorithm,
(2) we make no particular effort to modify our adversarial examples to improve
their chances of surviving the printing and photography process.
We simply make the scientific observation that very many adversarial examples
do survive this process without any intervention.
\citet{Sharif16AdvML} introduce extra features to make their attacks work
as best as possible for practical attacks against face recognition systems.
(3) \citet{Sharif16AdvML} are restricted in the number of pixels they can
modify (only those on the glasses frames) but can modify those pixels by
a large amount; we are restricted in the amount we can modify a pixel
but are free to modify all of them.

To investigate the extent to which adversarial examples survive in the
physical world,
we conducted an experiment with a pre-trained ImageNet Inception classifier~\citep{Inception-v3}.
We generated adversarial examples for this model,
then we fed these examples to the classifier through a cell-phone camera and measured the classification accuracy.
This scenario is a simple physical world system which perceives data through a camera and then runs image classification.
We found that a large fraction of adversarial examples generated for the original model remain misclassified
even when perceived through a camera.\footnote{
Dileep George noticed that another kind of adversarially constructed input,
designed to have no true class yet be categorized as belonging to a
specific class, fooled convolutional networks when photographed,
in a less systematic experiments.
As of August 19, 2016 it was mentioned in Figure 6 at~\url{http://www.evolvingai.org/fooling}
}

Surprisingly, our attack methodology required no modification to account for the
presence of the camera---the simplest possible attack of using adversarial
examples crafted for the Inception model resulted in adversarial examples that
successfully transferred to the union of the camera and Inception.
Our results thus provide a lower bound on the attack success rate that could
be achieved with more specialized attacks that explicitly model the camera
while crafting the adversarial example.

One limitation of our results is that we have assumed a threat model under which
the attacker has full knowledge of the model architecture and parameter values.
This is primarily so that we can use a single Inception v3 model in all experiments,
without having to devise and train a different high-performing model.
The adversarial example transfer property implies that our results could be extended
trivially to the scenario where the attacker does not have access to
the model description \citep{Szegedy-ICLR2014,Goodfellow-2015-adversarial,Papernot-2016-transferability}.
While we haven't run detailed experiments to study transferability of physical adversarial examples
we were able to build a simple phone application to demonstrate potential
adversarial black box attack in the physical world, see fig.~\ref{fig-phone-app-demo}.

To better understand how the non-trivial image transformations caused by the
camera affect adversarial example transferability,
we conducted a series of additional experiments where we studied how adversarial
examples transfer across several
specific kinds of synthetic image transformations.

The rest of the paper is structured as follows:
In Section~\ref{sec-adversarial-methods}, we review different methods
which we used to generate adversarial examples. This is followed in
Section~\ref{sec-adversarial-photos} by details about our ``physical world''
experimental set-up and results. Finally,
Section~\ref{sec-artificial-transform} describes our experiments with various
artificial image transformations (like changing brightness, contrast, etc...)
and how they affect adversarial examples.

\section{Methods of generating adversarial images}\label{sec-adversarial-methods}

This section describes different methods to generate adversarial examples which we have used in the experiments.
It is important to note that none of the described methods guarantees that generated image will be misclassified.
Nevertheless we call all of the generated images ``adversarial images''.

In the remaining of the paper we use the following notation:
\begin{itemize}
  \item $\bm{X}$ - an image, which is typically 3-D tensor (width $\times$ height $\times$ depth).
  In this paper, we assume that the values of the pixels are integer numbers in the range $[0, 255]$.
  \item $y_{true}$ - true class for the image $\bm{X}$.
  \item $J(\bm{X}, y)$ - cross-entropy cost function of the neural network, given image $\bm{X}$ and class $y$.
  We intentionally omit network weights (and other parameters) $\bm{\theta}$ in the cost function
  because we assume they are fixed (to the value resulting from training the machine learning model)
  in the context of the paper.
  For neural networks with a softmax output layer, the cross-entropy cost function applied to
  integer class labels equals the negative log-probability of the true class given the image:
  $J(\bm{X}, y) = - \log p(y | \bm{X})$, this relationship will be used below.
  \item $Clip_{X, \epsilon} \left\{ \bm{X}' \right\}$ - function which performs per-pixel
  clipping of the image $\bm{X}'$, so the result will be in $L_{\infty}$ $\epsilon$-neighbourhood of the source image $\bm{X}$.
  The exact clipping equation is as follows: 
  \[
  Clip_{X, \epsilon} \left\{ \bm{X}' \right\} (x, y, z) = \min \Bigl\{ 255, \bm{X}(x,y,z) + \epsilon, \max \bigl\{ 0, \bm{X}(x,y,z) - \epsilon, \bm{X}'(x,y,z) \bigr\} \Bigr\}\\
  \]
  where $\bm{X}(x,y,z)$ is the value of channel $z$ of the image $\bm{X}$ at coordinates $(x, y)$.
\end{itemize}

\subsection{Fast method}

One of the simplest methods to generate adversarial images, described in~\citep{Goodfellow-2015-adversarial},
is motivated by linearizing the cost function and solving for the perturbation
that maximizes the cost subject to an $L_{\infty}$ constraint.
This may be accomplished in closed form, for the cost of one call to back-propagation:
\[
\bm{X}^{adv} = \bm{X} + \epsilon \sign \bigl( \nabla_X J(\bm{X}, y_{true})  \bigr)
\]
where $\epsilon$ is a hyper-parameter to be chosen.

In this paper we refer to this method as ``fast'' because it does not require an iterative
procedure to compute adversarial examples, and thus is much faster than other considered methods.

\subsection{Basic iterative method}

We introduce a straightforward way to extend the ``fast'' method---we apply it multiple times with small step size,
and clip pixel values of intermediate results after each step to ensure that they are in an $\epsilon$-neighbourhood of the original image:

\[
\bm{X}^{adv}_{0} = \bm{X}, \quad
\bm{X}^{adv}_{N+1} = Clip_{X, \epsilon}\Bigl\{ \bm{X}^{adv}_{N} + \alpha \sign \bigl( \nabla_X J(\bm{X}^{adv}_{N}, y_{true})  \bigr) \Bigr\}
\]

In our experiments we used $\alpha = 1$, i.e. we changed the value of each pixel only by $1$ on each step.
We selected the number of iterations to be $\min(\epsilon + 4, 1.25\epsilon)$.
This amount of iterations was chosen heuristically; it is sufficient for the adversarial
example to reach the edge of the $\epsilon$ max-norm ball but restricted enough to
keep the computational cost of experiments manageable.

Below we refer to this method as ``basic iterative'' method.

\subsection{Iterative least-likely class method}

Both methods we have described so far simply try to increase the cost of the correct
class, without specifying which of the incorrect classes the model should select.
Such methods are sufficient for application to datasets such as MNIST and CIFAR-10,
where the number of classes is small and all classes are highly distinct from each
other.
On ImageNet, with a much larger number of classes and the varying degrees of significance
in the difference
between classes, these methods can result in uninteresting misclassifications,
such as mistaking one breed of sled dog for another breed of sled dog.
In order to create more interesting mistakes, we introduce the {\em iterative least-likely class method}.
This iterative method tries to make an adversarial image which will be classified as
a specific desired target class.
For desired class we chose the least-likely class according to the prediction of the trained network on image $\bm{X}$:
\[
y_{LL} = \argmin_{y} \bigl\{ p( y | \bm{X} ) \bigr\}.
\]
For a well-trained classifier, the least-likely class is usually highly dissimilar from
the true class, so this attack method results in more interesting mistakes, such
as mistaking a dog for an airplane.

To make an adversarial image which is classified as $y_{LL}$ we maximize $\log p(y_{LL} | \bm{X})$ by making iterative steps in the direction of
$\sign \bigl\{ \nabla_X \log p(y_{LL} | \bm{X}) \bigr\}$.
This last expression equals  $\sign \bigl\{ -\nabla_X J(\bm{X}, y_{LL}) \bigr)$ for neural networks with cross-entropy loss.
Thus we have the following procedure:

\[
\bm{X}^{adv}_{0} = \bm{X}, \quad
\bm{X}^{adv}_{N+1} = Clip_{X, \epsilon}\left\{ \bm{X}^{adv}_{N} - \alpha \sign \left( \nabla_X J(\bm{X}^{adv}_{N}, y_{LL})  \right) \right\}
\]

For this iterative procedure we used the same $\alpha$ and same number of iterations as for the basic iterative method.

Below we refer to this method as the ``least likely class'' method or shortly ``l.l. class''.

\subsection{Comparison of methods of generating adversarial examples}\label{sec-adv-methods-comparison}

\begin{figure}[!h]
  \centering
  \begin{subfigure}[b]{0.49\textwidth}
    \includegraphics[width=\textwidth]{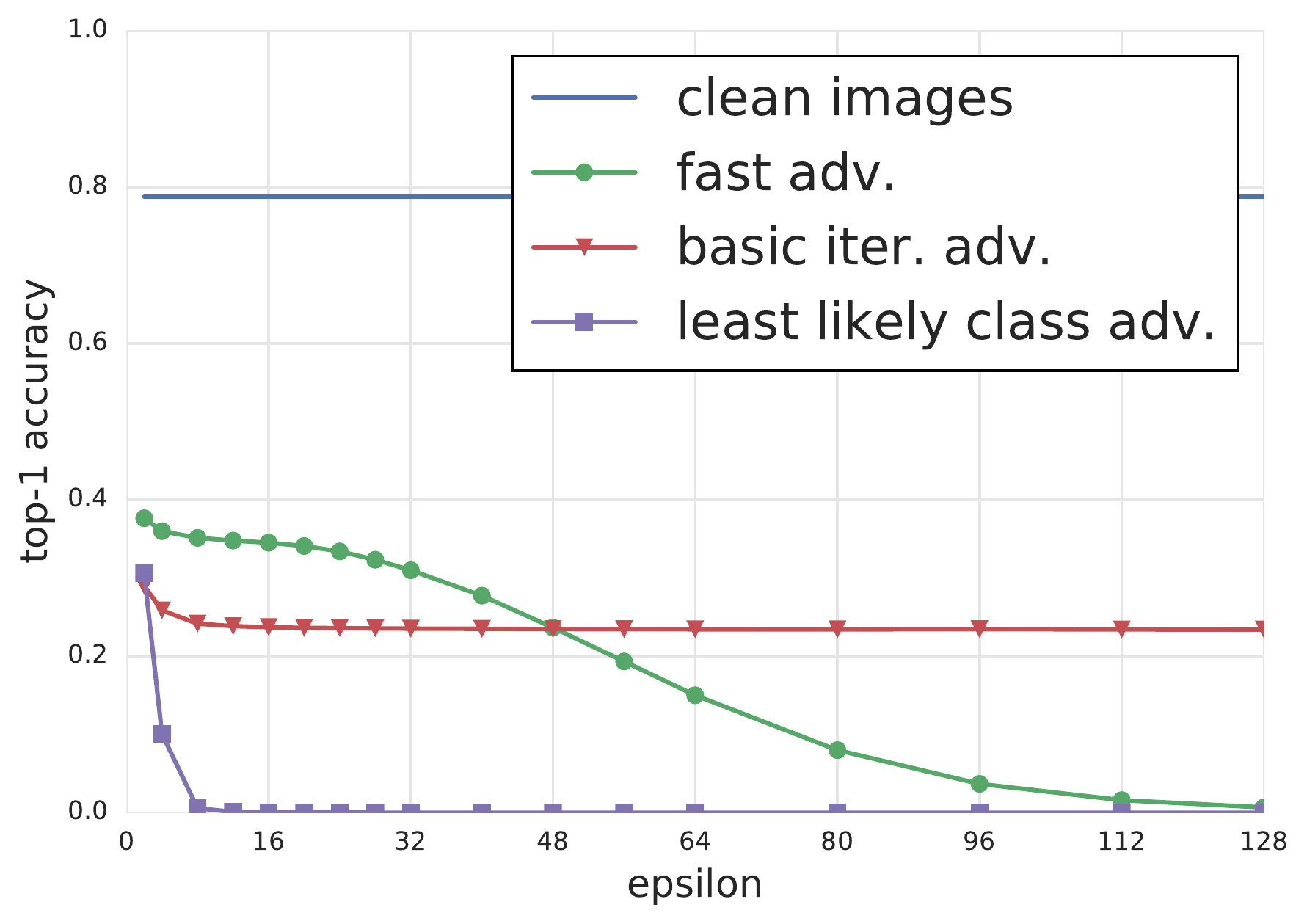}
  \end{subfigure}
  \begin{subfigure}[b]{0.49\textwidth}
    \includegraphics[width=\textwidth]{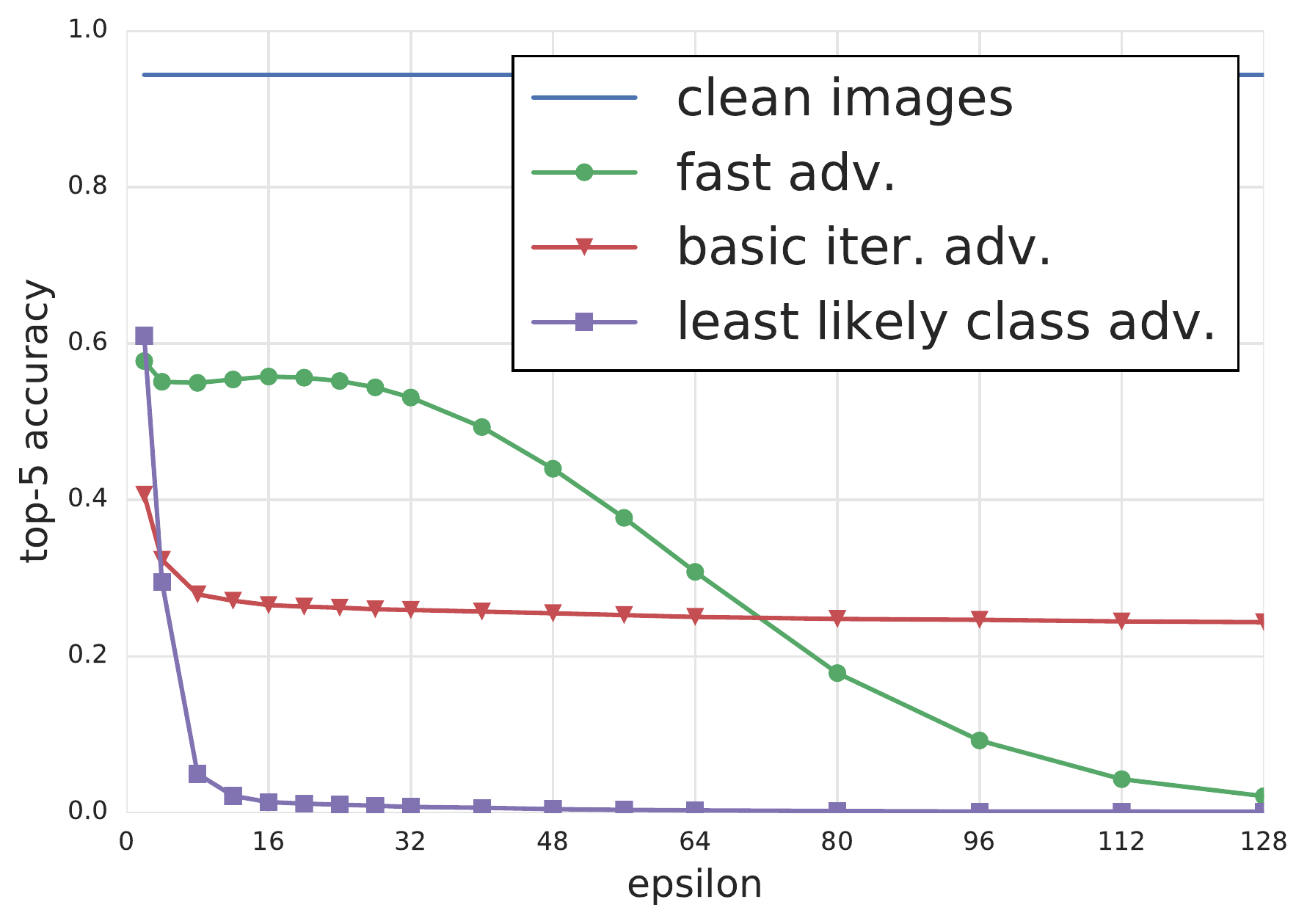}
  \end{subfigure}
  \caption{Top-1 and top-5 accuracy of Inception v3 under attack by
  different adversarial methods
  and different $\epsilon$ compared to ``clean images''~--- unmodified images from the dataset.
  The accuracy was computed on all $50,000$ validation images from the ImageNet dataset.
  In these experiments $\epsilon$ varies from $2$ to $128$.
  }\label{fig-acc-adversarial-methods}
\end{figure}

As mentioned above, it is not guaranteed that an adversarial image will actually
be misclassified---sometimes the attacker wins, and sometimes the machine learning
model wins.
We did an experimental comparison of adversarial methods to understand the actual classification accuracy on the generated 
images as well as the types of perturbations exploited by each of the methods.

The experiments were performed on all $50,000$ validation samples from the ImageNet dataset~\citep{russakovsky2014imagenet}
using a pre-trained Inception v3 classifier~\citep{Inception-v3}.
For each validation image, we generated adversarial examples using different methods and different values of $\epsilon$.
For each pair of method and $\epsilon$, we computed the classification accuracy on all $50,000$ images.
Also, we computed the accuracy on all clean images, which we used as a baseline.

Top-1 and top-5 classification accuracy on clean and adversarial images for various adversarial methods
are summarized in Figure~\ref{fig-acc-adversarial-methods}.
Examples of generated adversarial images could be found in Appendix in Figures~\ref{fig-adversarial-epsilon}~and~\ref{fig-adversarial-method}.

As shown in Figure~\ref{fig-acc-adversarial-methods}, the fast method decreases top-1 accuracy by a factor of two
and top-5 accuracy by about $40\%$ even with the smallest values of $\epsilon$.
As we increase $\epsilon$, accuracy on adversarial images generated by the fast method stays on approximately the same level
until $\epsilon=32$ and then slowly decreases to almost $0$ as $\epsilon$ grows to $128$.
This could be explained by the fact that the fast method adds $\epsilon$-scaled noise to each image,
thus higher values of $\epsilon$ essentially destroys the content of the image 
and makes it unrecognisable even by humans, see~Figure~\ref{fig-adversarial-epsilon}.

On the other hand iterative methods exploit much finer perturbations which do not destroy the image even with higher $\epsilon$
and at the same time confuse the classifier with higher rate.
The basic iterative method is able to produce better adversarial images when $\epsilon < 48$,
however as we increase $\epsilon$ it is unable to improve.
The ``least likely class'' method destroys the correct classification of most images even when $\epsilon$ is relatively small.

We limit all further experiments to $\epsilon \le 16$
because such perturbations are only perceived as a small noise (if perceived at all),
and adversarial methods are able to produce a significant number of misclassified examples
in this $\epsilon$-neighbourhood of clean images.

\section{Photos of adversarial examples}\label{sec-adversarial-photos}

\subsection{Destruction rate of adversarial images}\label{sec-destruction-rate}

To study the influence of arbitrary transformations on adversarial images we introduce the notion of destruction rate.
It can be described as the fraction of adversarial images which are no longer misclassified after the transformations.
The formal definition is the following:

\begin{equation}
\label{eq:destruction_rate}
d = \frac{ \sum_{k=1}^{n} C(\bm{X}^k, y_{true}^{k}) \overline{ C(\bm{X}_{adv}^k, y_{true}^{k}) } C(\Transform(\bm{X}_{adv}^k), y_{true}^{k})  }
{ \sum_{k=1}^{n} C(\bm{X}^k, y_{true}^{k}) \overline{ C(\bm{X}_{adv}^k, y_{true}^{k}) } }
\end{equation}

where $n$ is the number of images used to comput the destruction rate, $\bm{X}^k$ is an image from the dataset,
$y_{true}^{k}$ is the true class of this image,
and $\bm{X}_{adv}^k$ is the corresponding adversarial image.
The function $\Transform(\bullet)$ is an arbitrary image transformation---in this article,
we study a variety of transformations, including printing the image and taking a photo of
the result.
The function $C(\bm{X}, y)$ is an indicator function which returns whether the image was classified correctly:

\[
C(\bm{X}, y) =
\begin{cases}
  1, & \text{if image $\bm{X}$ is classified as $y$; } \\
  0, & \text{otherwise.}
\end{cases}
\]

We denote the binary negation of this indicator value as
$\overline{C(\bm{X}, y)}$, which is computed as $\overline{C(\bm{X}, y)} = 1 - C(\bm{X}, y)$.

\subsection{Experimental setup}

\begin{figure}[!ht]
  \centering
  \begin{subfigure}[b]{0.2\textwidth}
    \includegraphics[width=\textwidth]{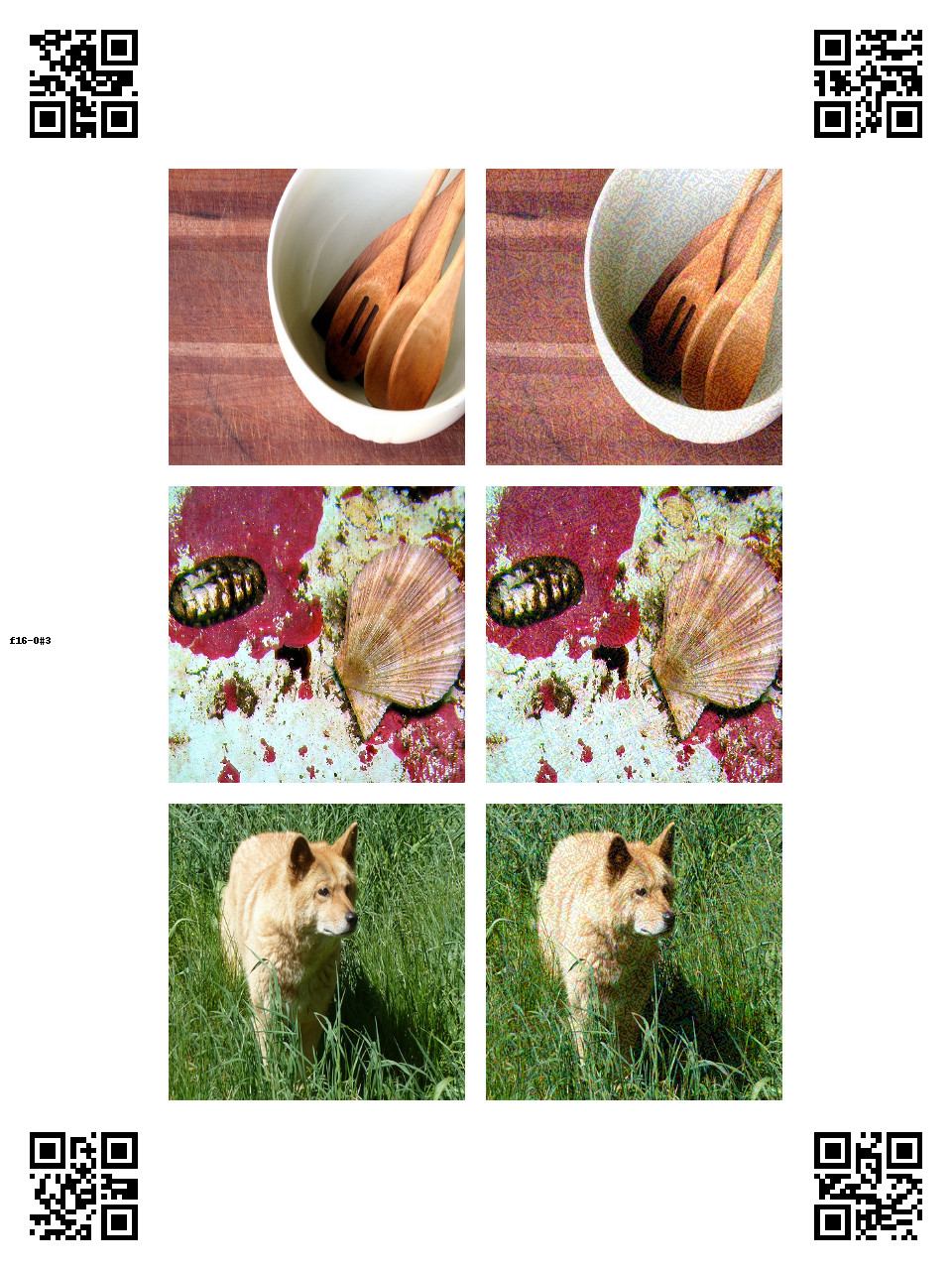}
    \caption{Printout}\label{pic-photos-setup-printout}
  \end{subfigure}
  \,
  \begin{subfigure}[b]{0.2\textwidth}
    \includegraphics[width=\textwidth]{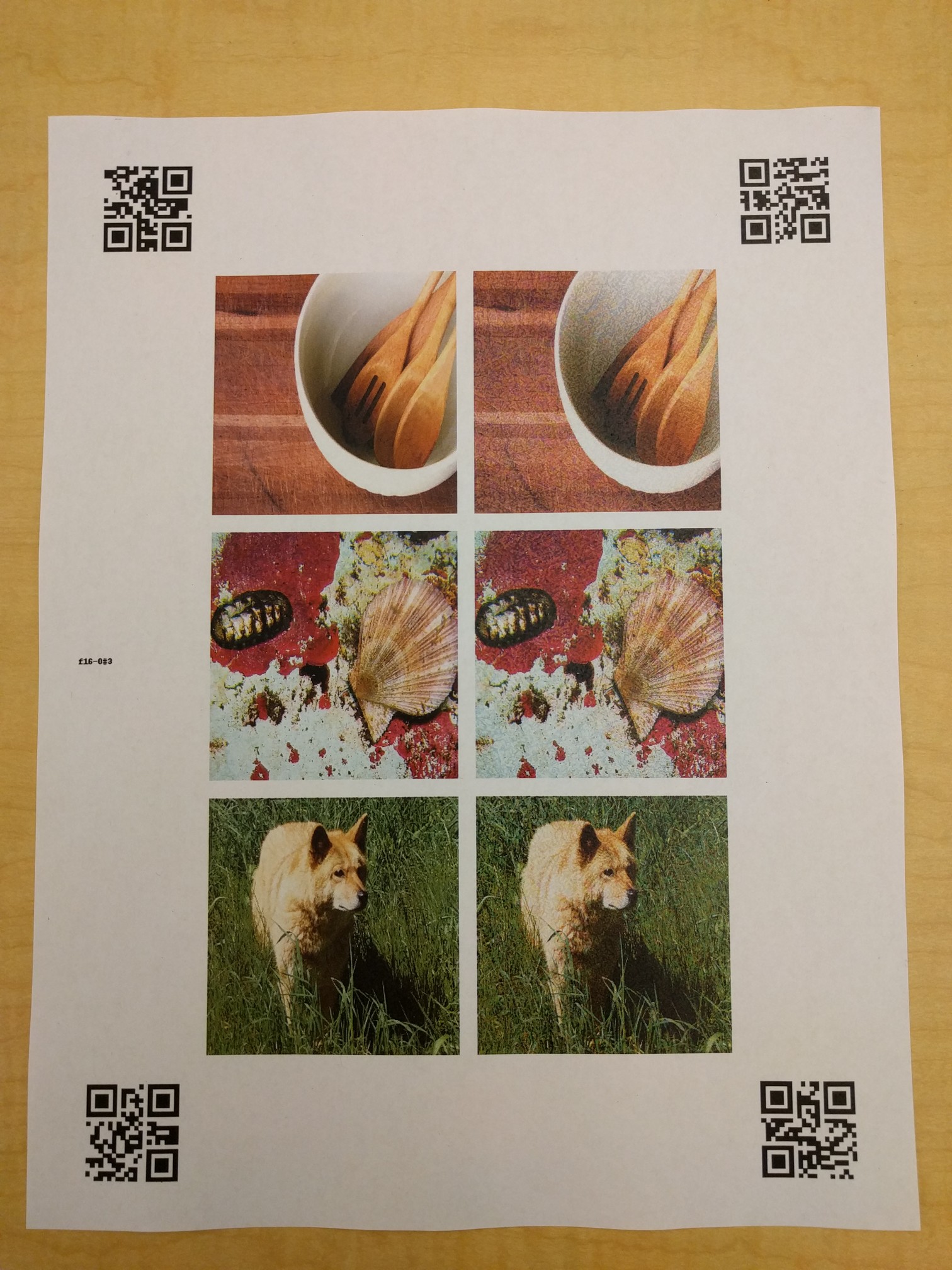}
    \caption{Photo of printout}\label{pic-photos-setup-photo}
  \end{subfigure}
  \,
  \begin{subfigure}[b]{0.2\textwidth}
    \includegraphics[width=\textwidth]{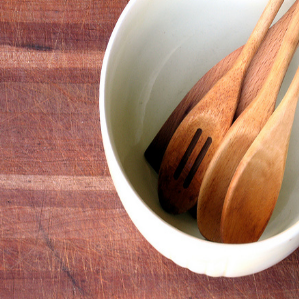}
    \caption{Cropped image}\label{pic-photos-setup-cropped}
  \end{subfigure}
  \caption{
  Experimental setup: (a) generated printout which contains pairs of clean and adversarial images,
  as well as QR codes to help automatic cropping;
  (b) photo of the printout made by a cellphone camera;
  (c) automatically cropped image from the photo.}\label{pic-photos-setup}
\end{figure}

To explore the possibility of physical adversarial examples we ran a series of experiments with photos of adversarial examples.
We printed clean and adversarial images, took photos of the printed pages, and cropped the printed images from
the photos of the full page.
We can think of this as a black box transformation that we refer to as ``photo transformation''.

We computed the accuracy on clean and adversarial images before and after
the photo transformation
as well as the destruction rate of adversarial images subjected to photo transformation.

The experimental procedure was as follows:

\begin{enumerate}
  \item Print the image, see Figure~\ref{pic-photos-setup-printout}.
  In order to reduce the amount of manual work, we printed multiple pairs of clean and adversarial examples on each sheet of paper.
  Also, QR codes were put into corners of the printout to facilitate automatic cropping.
  \begin{enumerate}
    \item All generated pictures of printouts~(Figure~\ref{pic-photos-setup-printout}) were saved in lossless PNG format.
    \item Batches of PNG printouts were converted to multi-page PDF file using the convert tool
          from the ImageMagick suite with the default settings: \texttt{convert *.png output.pdf}
    \item Generated PDF files were printed using a \textit{Ricoh MP C5503} office printer.
          Each page of PDF file was automatically scaled to fit the entire sheet of paper using the default printer scaling.
          The printer resolution was set to 600dpi.
  \end{enumerate}
  \item Take a photo of the printed image using a cell phone camera (Nexus 5x), see Figure~\ref{pic-photos-setup-photo}.
  \item Automatically crop and warp validation examples from the photo,
  so they would become squares of the same size as source images, see Figure~\ref{pic-photos-setup-cropped}:
  \begin{enumerate}
    \item Detect values and locations of four QR codes in the corners of the photo.
    The QR codes encode which batch of validation examples is shown on the photo.
    If detection of any of the corners failed, the entire photo was discarded and
    images from the photo were not used to calculate accuracy.
    We observed that no more than $10\%$ of all images were discarded in any experiment
    and typically the number of discarded images was about $3\%$ to $6\%$.
    \item Warp photo using perspective transform to move location of QR codes into pre-defined coordinates.
    \item After the image was warped, each example has known coordinates and can easily be cropped from the image.
  \end{enumerate}
  \item Run classification on transformed and source images. Compute accuracy and destruction rate of adversarial images.
\end{enumerate}

This procedure involves manually taking photos of the printed pages,
without careful control of lighting, camera angle, distance to the page, etc.
This is intentional; it introduces nuisance variability that has the potential
to destroy adversarial perturbations that depend on subtle, fine co-adaptation
of exact pixel values.
That being said, we did not intentionally seek out extreme camera angles or
lighting conditions. All photos were taken in normal indoor lighting with
the camera pointed approximately straight at the page.

For each combination of adversarial example generation method and $\epsilon$ we conducted two sets of experiments:

\begin{itemize}
  \item \textbf{Average case.}
  To measure the average case performance,
  we randomly selected 102 images to use in one experiment with a given $\epsilon$
  and adversarial method.
  This experiment estimates how often an adversary would succeed on randomly chosen
  photos---the world chooses an image randomly, and the adversary attempts
  to cause it to be misclassified.
  \item \textbf{Prefiltered case.}
  To study a more aggressive attack, we performed experiments in which the images
  are prefiltered.
  Specifically, we selected 102 images such that all clean images are classified
  correctly, and all adversarial images (before photo transformation) are classified incorrectly (both top-1 and top-5 classification).
  In addition we used a confidence threshold for the top prediction: $p(y_{predicted} | \bm{X}) \ge 0.8$,
  where $y_{predicted}$ is the class predicted by the network for image $\bm{X}$.
  This experiment measures how often an adversary would succeed when the adversary can choose
  the original image to attack.
  Under our threat model, the adversary has access to the model parameters and architecture,
  so the attacker can always run inference to determine whether an attack will succeed in
  the absence of photo transformation.
  The attacker might expect to do the best by choosing to make attacks that succeed in this
  initial condition.
  The victim then takes a new photo of the physical object that the attacker chooses to display,
  and the photo transformation can either preserve the attack or destroy it.
\end{itemize}

\subsection{Experimental results on photos of adversarial images}

\begin{table}[!h]
\caption{Accuracy on photos of adversarial images in the average case (randomly chosen images).}
\label{table-acc-photos-average}
\begin{center}
\begin{tabular}{|l||c|c|c|c||c|c|c|c|}
\hline
                       & \multicolumn{4}{|c||}{Photos} & \multicolumn{4}{|c|}{Source images} \\ \cline{2-9}
Adversarial & \multicolumn{2}{|c|}{Clean images} & \multicolumn{2}{|c||}{Adv. images} &
                           \multicolumn{2}{|c|}{Clean images} & \multicolumn{2}{|c|}{Adv. images} \\ \cline{2-9}
method       & top-1 & top-5 & top-1 & top-5 & top-1 & top-5 & top-1 & top-5 \\
\hline
fast $\epsilon=16$ & 79.8\% & 91.9\% & 36.4\% & 67.7\% & 85.3\% & 94.1\% & 36.3\% & 58.8\% \\
fast $\epsilon=8$ & 70.6\% & 93.1\% & 49.0\% & 73.5\% & 77.5\% & 97.1\% & 30.4\% & 57.8\% \\
fast $\epsilon=4$ & 72.5\% & 90.2\% & 52.9\% & 79.4\% & 77.5\% & 94.1\% & 33.3\% & 51.0\% \\
fast $\epsilon=2$ & 65.7\% & 85.9\% & 54.5\% & 78.8\% & 71.6\% & 93.1\% & 35.3\% & 53.9\% \\
iter. basic $\epsilon=16$ & 72.9\% & 89.6\% & 49.0\% & 75.0\% & 81.4\% & 95.1\% & 28.4\% & 31.4\% \\
iter. basic $\epsilon=8$ & 72.5\% & 93.1\% & 51.0\% & 87.3\% & 73.5\% & 93.1\% & 26.5\% & 31.4\% \\
iter. basic $\epsilon=4$ & 63.7\% & 87.3\% & 48.0\% & 80.4\% & 74.5\% & 92.2\% & 12.7\% & 24.5\% \\
iter. basic $\epsilon=2$ & 70.7\% & 87.9\% & 62.6\% & 86.9\% & 74.5\% & 96.1\% & 28.4\% & 41.2\% \\
l.l. class $\epsilon=16$ & 71.1\% & 90.0\% & 60.0\% & 83.3\% & 79.4\% & 96.1\% & 1.0\% & 1.0\% \\
l.l. class $\epsilon=8$ & 76.5\% & 94.1\% & 69.6\% & 92.2\% & 78.4\% & 98.0\% & 0.0\% & 6.9\% \\
l.l. class $\epsilon=4$ & 76.8\% & 86.9\% & 75.8\% & 85.9\% & 80.4\% & 90.2\% & 9.8\% & 24.5\% \\
l.l. class $\epsilon=2$ & 71.6\% & 87.3\% & 68.6\% & 89.2\% & 75.5\% & 92.2\% & 20.6\% & 44.1\% \\ 
\hline
\end{tabular}
\end{center}
\end{table}

\begin{table}[!h]
\caption{Accuracy on photos of adversarial images in the prefiltered case
(clean image correctly classified, adversarial image confidently incorrectly classified
in digital form being being printed and photographed
).}
\label{table-acc-photos-worst}
\begin{center}
\begin{tabular}{|l||c|c|c|c||c|c|c|c|}
\hline
                       & \multicolumn{4}{|c||}{Photos} & \multicolumn{4}{|c|}{Source images} \\ \cline{2-9}
Adversarial & \multicolumn{2}{|c|}{Clean images} & \multicolumn{2}{|c||}{Adv. images} &
                           \multicolumn{2}{|c|}{Clean images} & \multicolumn{2}{|c|}{Adv. images} \\ \cline{2-9}
method       & top-1 & top-5 & top-1 & top-5 & top-1 & top-5 & top-1 & top-5 \\
\hline
fast $\epsilon=16$ & 81.8\% & 97.0\% & 5.1\% & 39.4\% & 100.0\% & 100.0\% & 0.0\% & 0.0\% \\
fast $\epsilon=8$ & 77.1\% & 95.8\% & 14.6\% & 70.8\% & 100.0\% & 100.0\% & 0.0\% & 0.0\% \\
fast $\epsilon=4$ & 81.4\% & 100.0\% & 32.4\% & 91.2\% & 100.0\% & 100.0\% & 0.0\% & 0.0\% \\
fast $\epsilon=2$ & 88.9\% & 99.0\% & 49.5\% & 91.9\% & 100.0\% & 100.0\% & 0.0\% & 0.0\% \\
iter. basic $\epsilon=16$ & 93.3\% & 97.8\% & 60.0\% & 87.8\% & 100.0\% & 100.0\% & 0.0\% & 0.0\% \\
iter. basic $\epsilon=8$ & 89.2\% & 98.0\% & 64.7\% & 91.2\% & 100.0\% & 100.0\% & 0.0\% & 0.0\% \\
iter. basic $\epsilon=4$ & 92.2\% & 97.1\% & 77.5\% & 94.1\% & 100.0\% & 100.0\% & 0.0\% & 0.0\% \\
iter. basic $\epsilon=2$ & 93.9\% & 97.0\% & 80.8\% & 97.0\% & 100.0\% & 100.0\% & 0.0\% & 1.0\% \\
l.l. class $\epsilon=16$ & 95.8\% & 100.0\% & 87.5\% & 97.9\% & 100.0\% & 100.0\% & 0.0\% & 0.0\% \\
l.l. class $\epsilon=8$ & 96.0\% & 100.0\% & 88.9\% & 97.0\% & 100.0\% & 100.0\% & 0.0\% & 0.0\% \\
l.l. class $\epsilon=4$ & 93.9\% & 100.0\% & 91.9\% & 98.0\% & 100.0\% & 100.0\% & 0.0\% & 0.0\% \\
l.l. class $\epsilon=2$ & 92.2\% & 99.0\% & 93.1\% & 98.0\% & 100.0\% & 100.0\% & 0.0\% & 0.0\% \\ 
\hline
\end{tabular}
\end{center}
\end{table}

\begin{table}[!h]
\caption{Adversarial image destruction rate with photos.}
\label{table-destruction-rate}
\begin{center}
\begin{tabular}{|l||c|c||c|c|}
\hline
Adversarial & \multicolumn{2}{|c||}{Average case} & \multicolumn{2}{|c|}{Prefiltered case}  \\ \cline{2-5}
method       & top-1 & top-5 & top-1 & top-5 \\ 
\hline
fast $\epsilon=16$ & 12.5\% & 40.0\% & 5.1\% & 39.4\% \\
fast $\epsilon=8$ & 33.3\% & 40.0\% & 14.6\% & 70.8\% \\
fast $\epsilon=4$ & 46.7\% & 65.9\% & 32.4\% & 91.2\% \\
fast $\epsilon=2$ & 61.1\% & 63.2\% & 49.5\% & 91.9\% \\
iter. basic $\epsilon=16$ & 40.4\% & 69.4\% & 60.0\% & 87.8\% \\
iter. basic $\epsilon=8$ & 52.1\% & 90.5\% & 64.7\% & 91.2\% \\
iter. basic $\epsilon=4$ & 52.4\% & 82.6\% & 77.5\% & 94.1\% \\
iter. basic $\epsilon=2$ & 71.7\% & 81.5\% & 80.8\% & 96.9\% \\
l.l. class $\epsilon=16$ & 72.2\% & 85.1\% & 87.5\% & 97.9\% \\
l.l. class $\epsilon=8$ & 86.3\% & 94.6\% & 88.9\% & 97.0\% \\
l.l. class $\epsilon=4$ & 90.3\% & 93.9\% & 91.9\% & 98.0\% \\
l.l. class $\epsilon=2$ & 82.1\% & 93.9\% & 93.1\% & 98.0\% \\
\hline
\end{tabular}
\end{center}
\end{table}

Results of the photo transformation experiment are summarized in Tables~\ref{table-acc-photos-average}, \ref{table-acc-photos-worst} and~\ref{table-destruction-rate}.

We found that ``fast'' adversarial images are more robust to photo transformation compared to iterative methods.
This could be explained by the fact that iterative methods exploit more subtle kind of perturbations,
and these subtle perturbations are more likely to be destroyed by photo transformation.

One unexpected result is that in some cases the adversarial destruction rate in
the ``prefiltered case'' was higher compared to the ``average case''.
In the case of the iterative methods, even the total success rate was lower
for prefiltered images rather than randomly selected images.
This suggests that, to obtain very high confidence, iterative methods often
make subtle co-adaptations that are not able to survive photo transformation.

Overall, the results show that some fraction of adversarial examples stays
misclassified even after a non-trivial transformation: the photo transformation.
This demonstrates the possibility of physical adversarial examples.
For example, an adversary using the fast method with $\epsilon=16$
could expect that about $2/3$ of the images would be top-1 misclassified
and about $1/3$ of the images would be top-5 misclassified.
Thus by generating enough adversarial images, the adversary could expect to
cause far more misclassification than would occur on natural inputs.

\subsection{Demonstration of black box adversarial attack in the physical world}

The experiments described above study physical adversarial examples under the assumption
that adversary has full access to the model (i.e. the adversary knows the architecture, model weights, etc \ldots).
However, the black box scenario, in which the attacker does not have access to the
model, is a more realistic model of many security threats.
Because adversarial examples often transfer from one model to another, they may
be used for black box attacks \citet{Szegedy-ICLR2014,Papernot-MGJCS16}.
As our own black box attack, we demonstrated that our physical adversarial examples
fool a different model than the one that was used to construct them.
Specifically, we showed that they fool the 
open source TensorFlow camera demo~\footnote{
As of October 25, 2016 TensorFlow camera demo was available
at~\url{https://github.com/tensorflow/tensorflow/tree/master/tensorflow/examples/android}}
--- an app for mobile phones which performs image classification on-device.
We showed several printed clean and adversarial images to this app and observed change of classification
from true label to incorrect label.
Video with the demo available at~\url{https://youtu.be/zQ_uMenoBCk}.
We also demonstrated this effect live at GeekPwn 2016.

\section{Artificial image transformations}\label{sec-artificial-transform}

The transformations applied to images by the process of printing them, photographing them, and cropping
them 
could be considered as some combination of much simpler image transformations.
Thus to better understand what is going on we conducted a series of experiments to measure the adversarial
destruction rate on artificial image transformations.
We explored the following set of transformations: change of contrast and brightness, Gaussian blur, Gaussian noise, and JPEG encoding.

For this set of experiments we used a subset of $1,000$ images randomly selected from the validation set.
This subset of $1,000$ images was selected once, thus all experiments from this section used the same subset of images.
We performed experiments for multiple pairs of adversarial method and transformation.
For each given pair of transformation and adversarial method we computed adversarial examples,
applied the transformation to the adversarial examples,
and then computed the destruction rate according to Equation~(\ref{eq:destruction_rate}).

Detailed results for various transformations and adversarial methods with $\epsilon = 16$
could be found in Appendix in Figure~\ref{fig-transform-destruction-rate}.
The following general observations can be drawn from these experiments:
\begin{itemize}
  \item Adversarial examples generated by the fast method are the most robust to transformations,
  and adversarial examples generated by the iterative least-likely class method are the least robust.
  This coincides with our results on photo transformation.
  \item The top-5 destruction rate is typically higher than top-1 destruction rate.
  This can be explained by the fact that in order to ``destroy'' top-5 adversarial examples, a transformation has to push the
  correct class labels into one of the top-5 predictions.
  However in order to destroy top-1 adversarial examples we have to push the
  correct label to be top-1 prediction, which is a strictly stronger requirement.
  \item Changing brightness and contrast does not affect adversarial examples much.
  The destruction rate on fast and basic iterative adversarial examples is less
  than $5\%$, and for the iterative least-likely class method it is less than $20\%$.
  \item Blur, noise and JPEG encoding have a higher destruction rate than
  changes of brightness and contrast.
  In particular, the destruction rate for iterative methods could reach $80\%-90\%$.
  However none of these transformations destroy $100\%$ of adversarial examples,
  which coincides with the ``photo transformation'' experiment.
\end{itemize}

\section{Conclusion}

In this paper we explored the possibility of creating adversarial examples for
machine learning systems which operate in the physical world.
We used images taken from a cell-phone camera as an input to an Inception v3 image classification neural network.
We showed that in such a set-up, a significant fraction of adversarial images crafted using the original network
are misclassified even when fed to the classifier through the camera.
This finding demonstrates the possibility of adversarial examples for machine learning systems in the physical world.
In future work, we expect that it will be possible to demonstrate attacks
using other kinds of physical objects besides images printed on paper,
attacks against different kinds of machine learning systems,
such as sophisticated reinforcement learning agents,
attacks performed without access to the model's parameters and architecture (presumably using the transfer property),
and physical attacks that achieve a higher success rate by explicitly modeling
the phyiscal transformation during the adversarial example construction process.
We also hope that future work will develop effective methods for defending against
such attacks.

\bibliography{ml,adversarial_photos}
\bibliographystyle{iclr2017_workshop}

\newpage
\begin{appendices}

Appendix contains following figures:
\begin{enumerate}
\item Figure~\ref{fig-adversarial-method} with examples of adversarial images produced by different adversarial methods.
\item Figure~\ref{fig-adversarial-epsilon} with examples of adversarial images for various values of $\epsilon$.
\item Figure~\ref{fig-transform-destruction-rate} contain plots of adversarial destruction rates for various image transformations.
\end{enumerate}

\begin{figure}[h]
  \captionsetup[subfigure]{labelformat=empty}
  \centering
  \begin{subfigure}[b]{0.49\textwidth}
    \includegraphics[width=\textwidth]{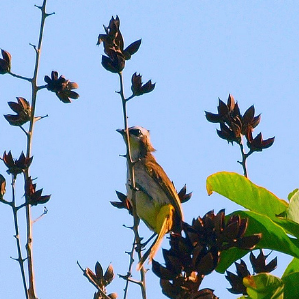}
    \caption{Clean image}
  \end{subfigure}
  \begin{subfigure}[b]{0.49\textwidth}
    \includegraphics[width=\textwidth]{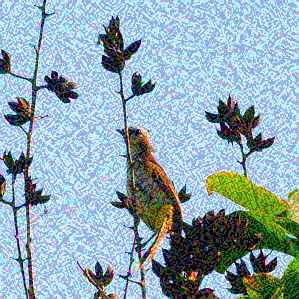}
    \caption{``Fast''; $L_{\infty}$ distance to clean image = $32$}
  \end{subfigure}
  \begin{subfigure}[b]{0.49\textwidth}
    \includegraphics[width=\textwidth]{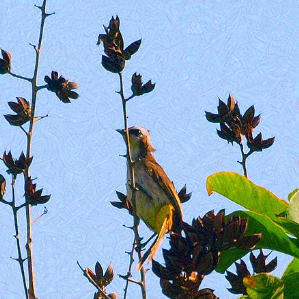}
    \caption{``Basic iter.''; $L_{\infty}$ distance to clean image = $32$}
  \end{subfigure}
  \begin{subfigure}[b]{0.49\textwidth}
    \includegraphics[width=\textwidth]{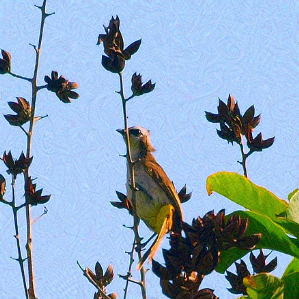}
    \caption{``L.l. class''; $L_{\infty}$ distance to clean image = $28$}
  \end{subfigure}
  \caption{Comparison of different adversarial methods with $\epsilon = 32$.
  Perturbations generated by iterative methods are finer compared to the fast method.
  Also iterative methods do not always select a point on the border of $\epsilon$-neighbourhood
  as an adversarial image.
  }\label{fig-adversarial-method}
\end{figure}

\begin{figure}[h]
  \captionsetup[subfigure]{labelformat=empty}
  \centering
  \begin{subfigure}[b]{0.24\textwidth}
    \includegraphics[width=\textwidth]{img0413}
    \caption{clean image}
  \end{subfigure}
  \begin{subfigure}[b]{0.24\textwidth}
    \includegraphics[width=\textwidth]{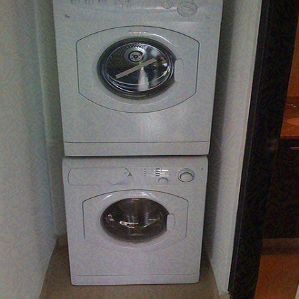}
    \caption{$\epsilon = 4$}
  \end{subfigure}
  \begin{subfigure}[b]{0.24\textwidth}
    \includegraphics[width=\textwidth]{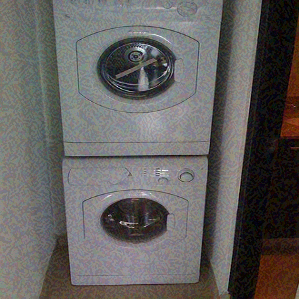}
    \caption{$\epsilon = 8$}
  \end{subfigure}
  \begin{subfigure}[b]{0.24\textwidth}
    \includegraphics[width=\textwidth]{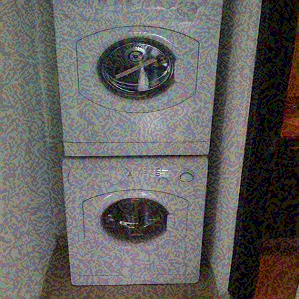}
    \caption{$\epsilon = 16$}
  \end{subfigure}
  \\
  \begin{subfigure}[b]{0.24\textwidth}
    \includegraphics[width=\textwidth]{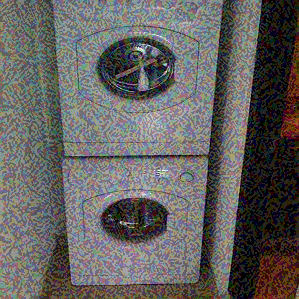}
    \caption{$\epsilon = 24$}
  \end{subfigure}
  \begin{subfigure}[b]{0.24\textwidth}
    \includegraphics[width=\textwidth]{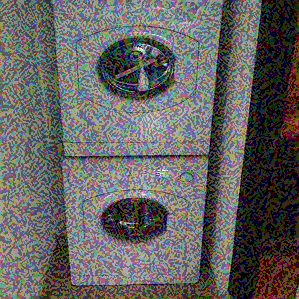}
    \caption{$\epsilon = 32$}
  \end{subfigure}
  \begin{subfigure}[b]{0.24\textwidth}
    \includegraphics[width=\textwidth]{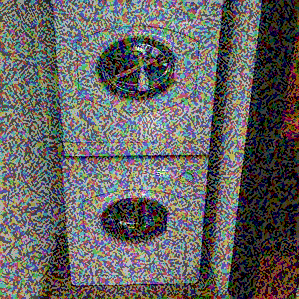}
    \caption{$\epsilon = 48$}
  \end{subfigure}
  \begin{subfigure}[b]{0.24\textwidth}
    \includegraphics[width=\textwidth]{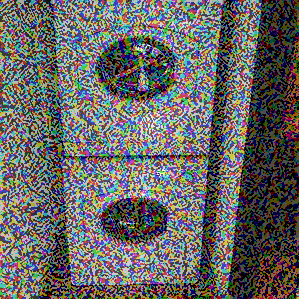}
    \caption{$\epsilon = 64$}
  \end{subfigure}
  \\
  \begin{subfigure}[b]{0.24\textwidth}
    \includegraphics[width=\textwidth]{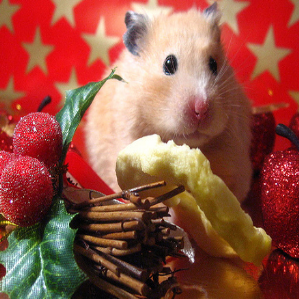}
    \caption{clean image}
  \end{subfigure}
  \begin{subfigure}[b]{0.24\textwidth}
    \includegraphics[width=\textwidth]{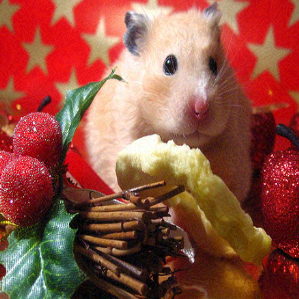}
    \caption{$\epsilon = 4$}
  \end{subfigure}
  \begin{subfigure}[b]{0.24\textwidth}
    \includegraphics[width=\textwidth]{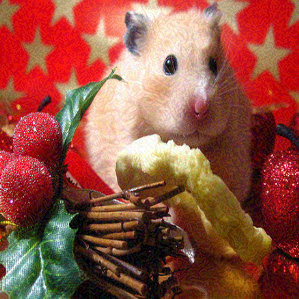}
    \caption{$\epsilon = 8$}
  \end{subfigure}
  \begin{subfigure}[b]{0.24\textwidth}
    \includegraphics[width=\textwidth]{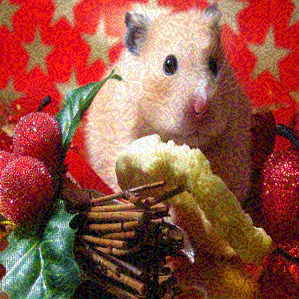}
    \caption{$\epsilon = 16$}
  \end{subfigure}
  \\
  \begin{subfigure}[b]{0.24\textwidth}
    \includegraphics[width=\textwidth]{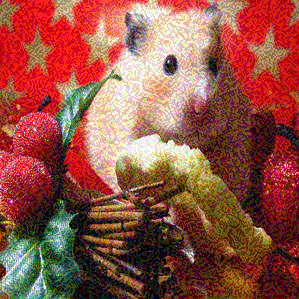}
    \caption{$\epsilon = 24$}
  \end{subfigure}
  \begin{subfigure}[b]{0.24\textwidth}
    \includegraphics[width=\textwidth]{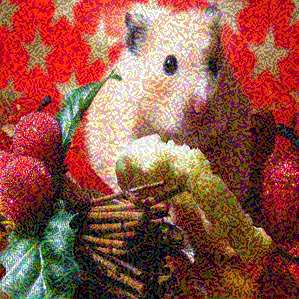}
    \caption{$\epsilon = 32$}
  \end{subfigure}
  \begin{subfigure}[b]{0.24\textwidth}
    \includegraphics[width=\textwidth]{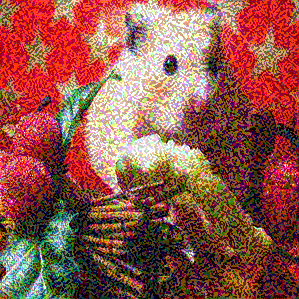}
    \caption{$\epsilon = 48$}
  \end{subfigure}
  \begin{subfigure}[b]{0.24\textwidth}
    \includegraphics[width=\textwidth]{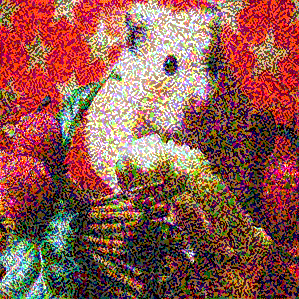}
    \caption{$\epsilon = 64$}
  \end{subfigure}
  \caption{Comparison of images resulting from an adversarial pertubation using the ``fast'' method with
  different size of perturbation $\epsilon$.
  The top image is a ``washer'' while the bottom one is a ``hamster''.
  In both cases clean images are classified correctly and
  adversarial images are misclassified for all considered $\epsilon$.
  }\label{fig-adversarial-epsilon}
\end{figure}

\begin{figure}[h]
  \centering
  \begin{subfigure}[b]{0.49\textwidth}
    \includegraphics[width=\textwidth]{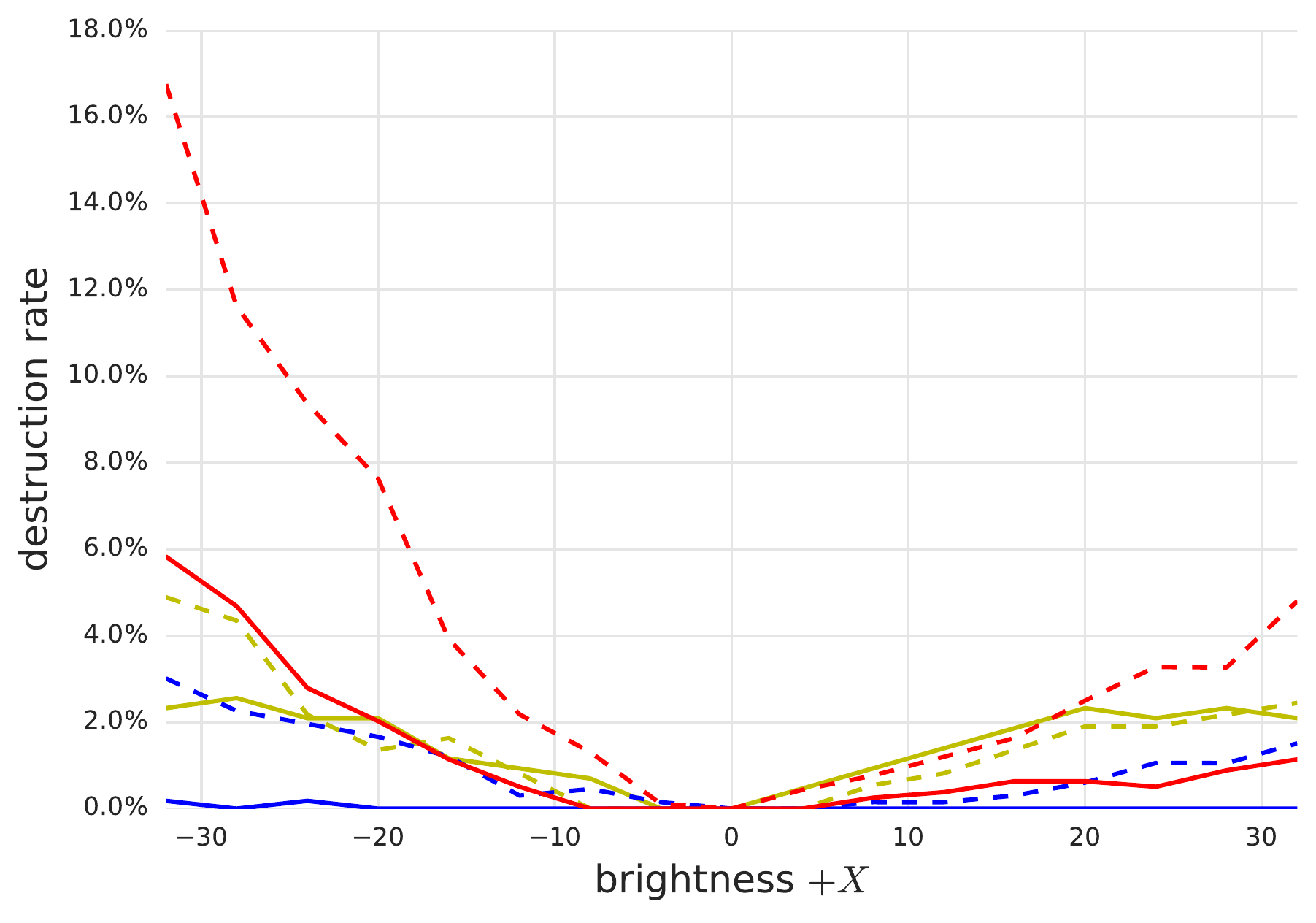}
    \caption{Change of brightness}
  \end{subfigure}
  \begin{subfigure}[b]{0.49\textwidth}
    \includegraphics[width=\textwidth]{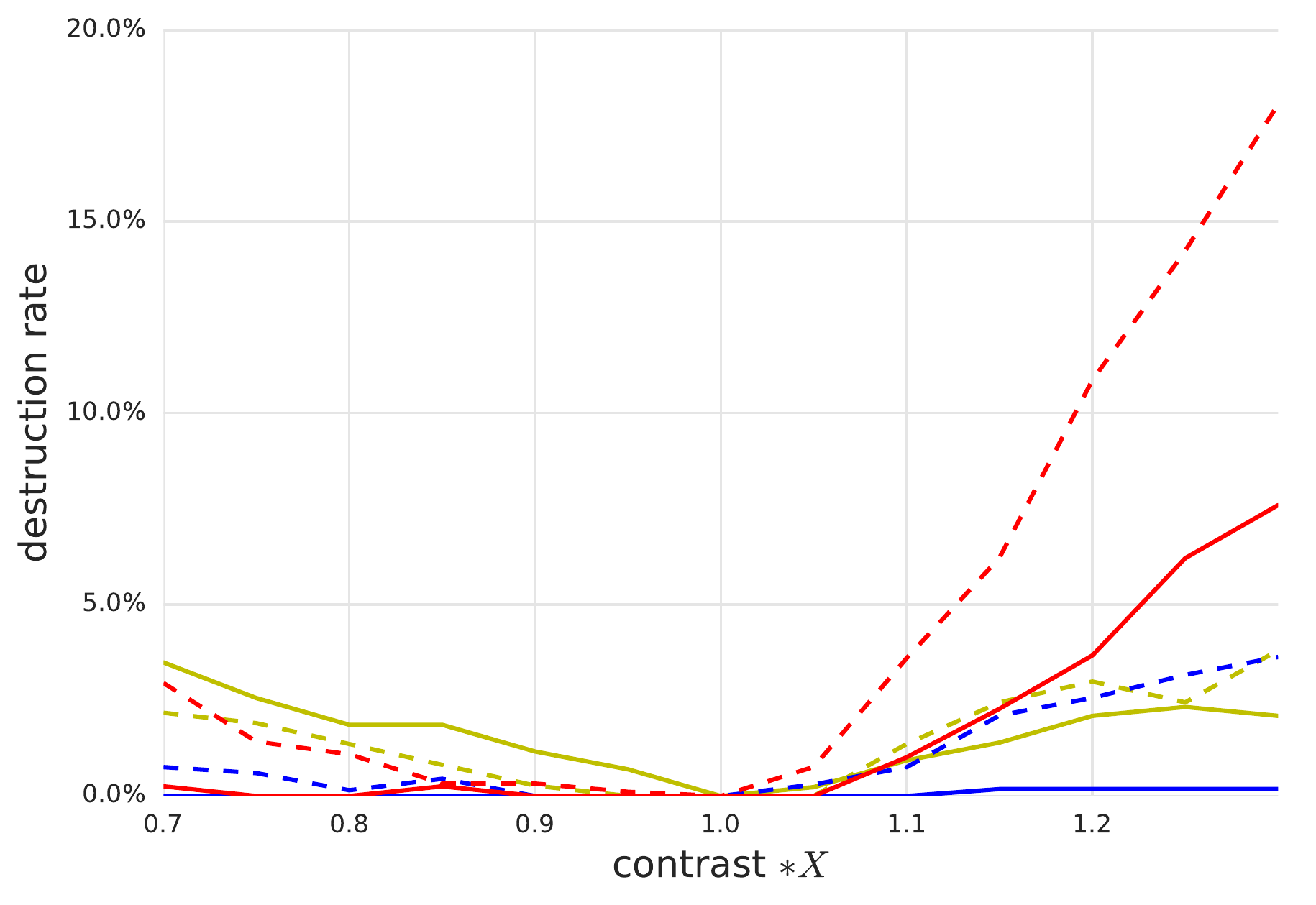}
    \caption{Change of contrast}
  \end{subfigure}
  \\
  \begin{subfigure}[b]{0.49\textwidth}
    \includegraphics[width=\textwidth]{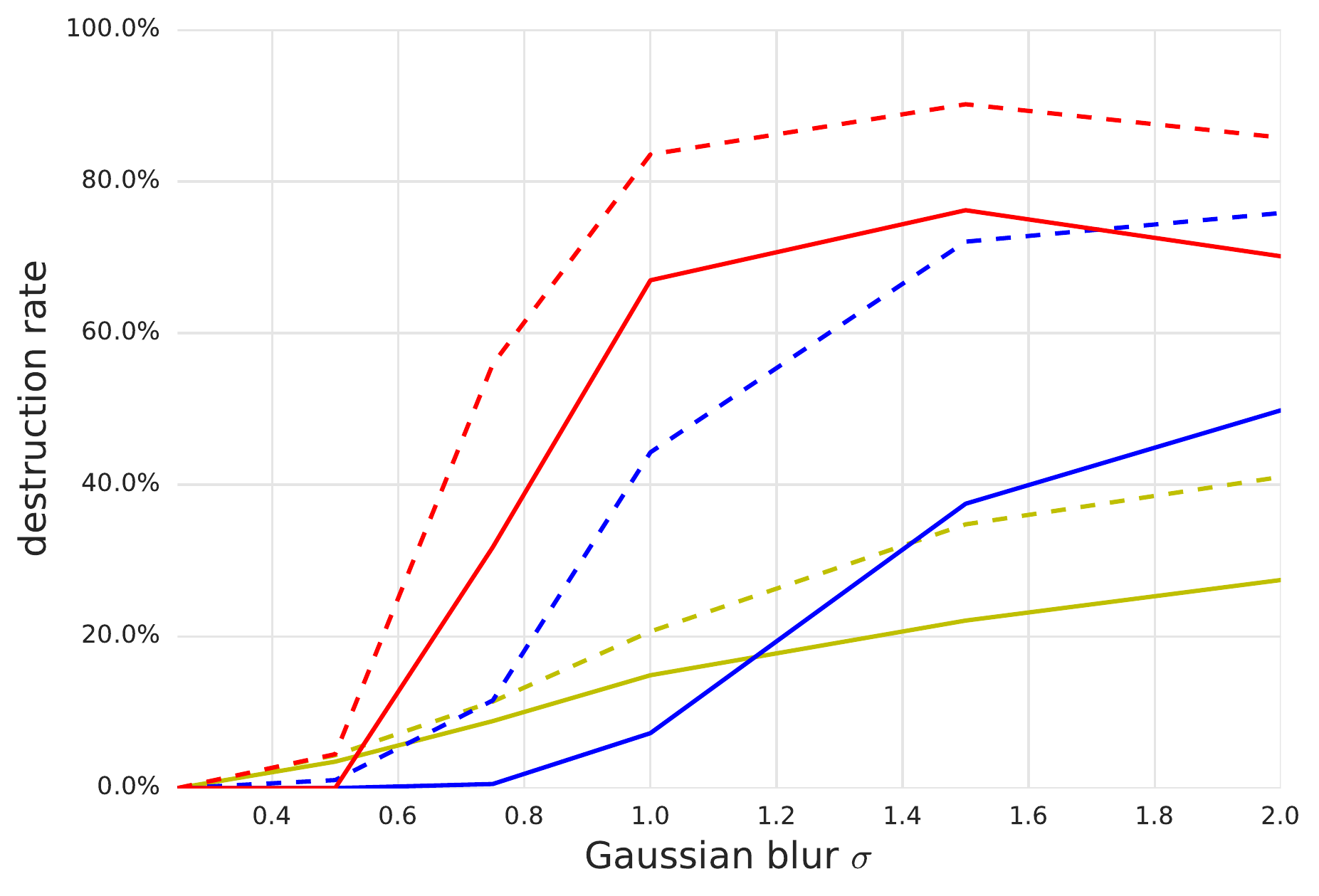}
    \caption{Gaussian blur}
  \end{subfigure}
  \begin{subfigure}[b]{0.49\textwidth}
    \includegraphics[width=\textwidth]{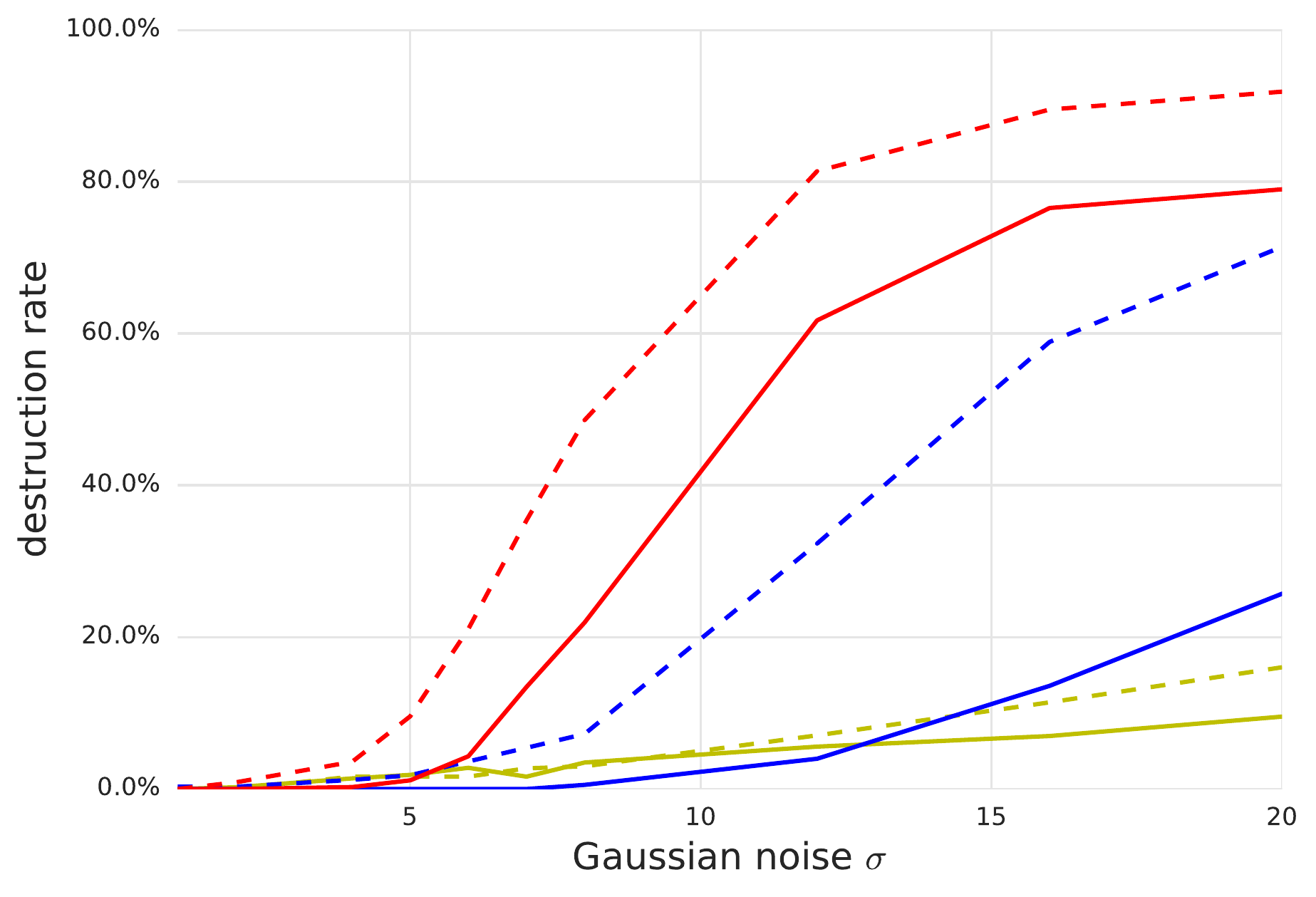}
    \caption{Gaussian noise}
  \end{subfigure}
  \\
  \begin{subfigure}[b]{0.49\textwidth}
    \includegraphics[width=\textwidth]{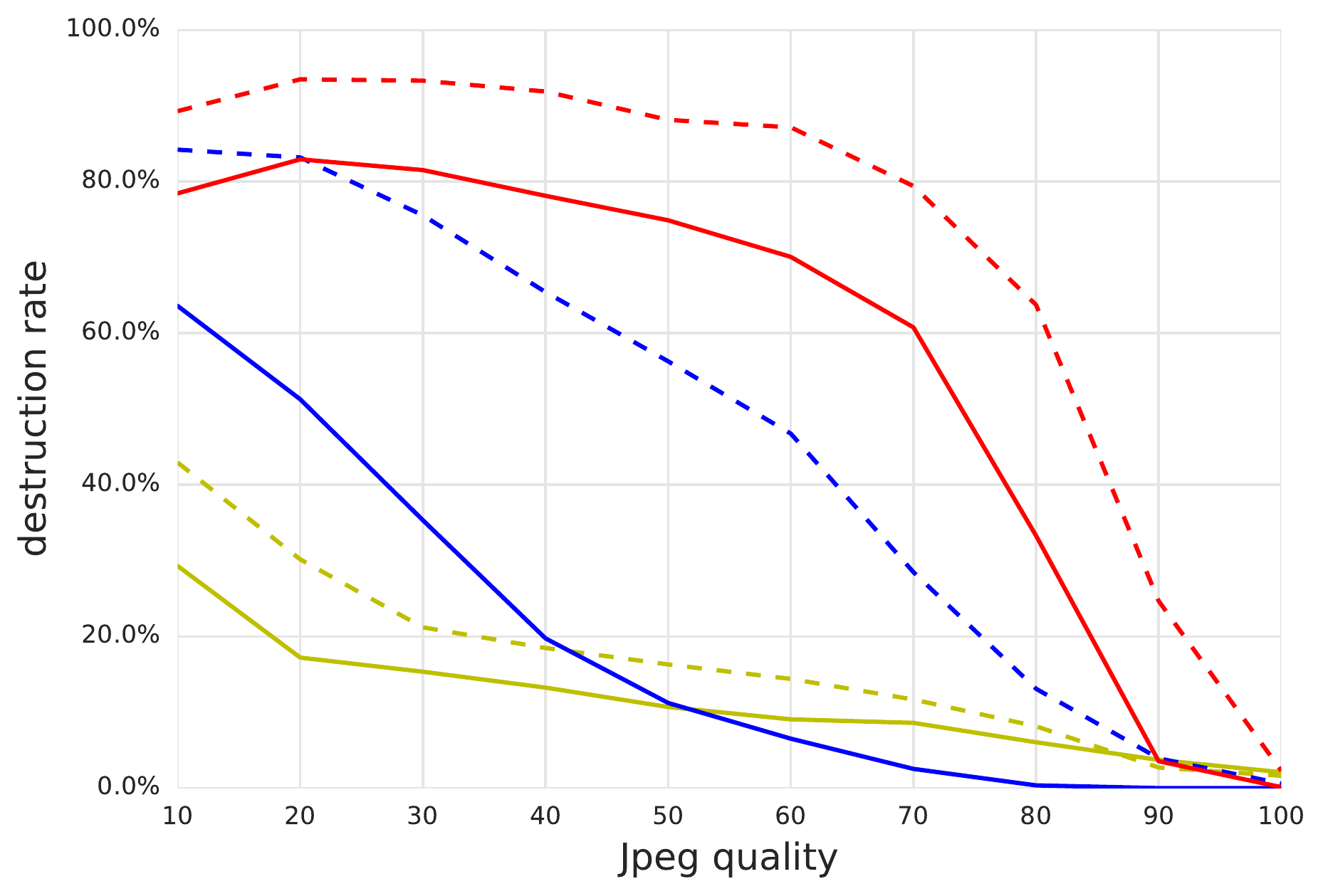}
    \caption{JPEG encoding}
  \end{subfigure}
  \begin{subfigure}[b]{0.49\textwidth}
    \includegraphics[width=\textwidth]{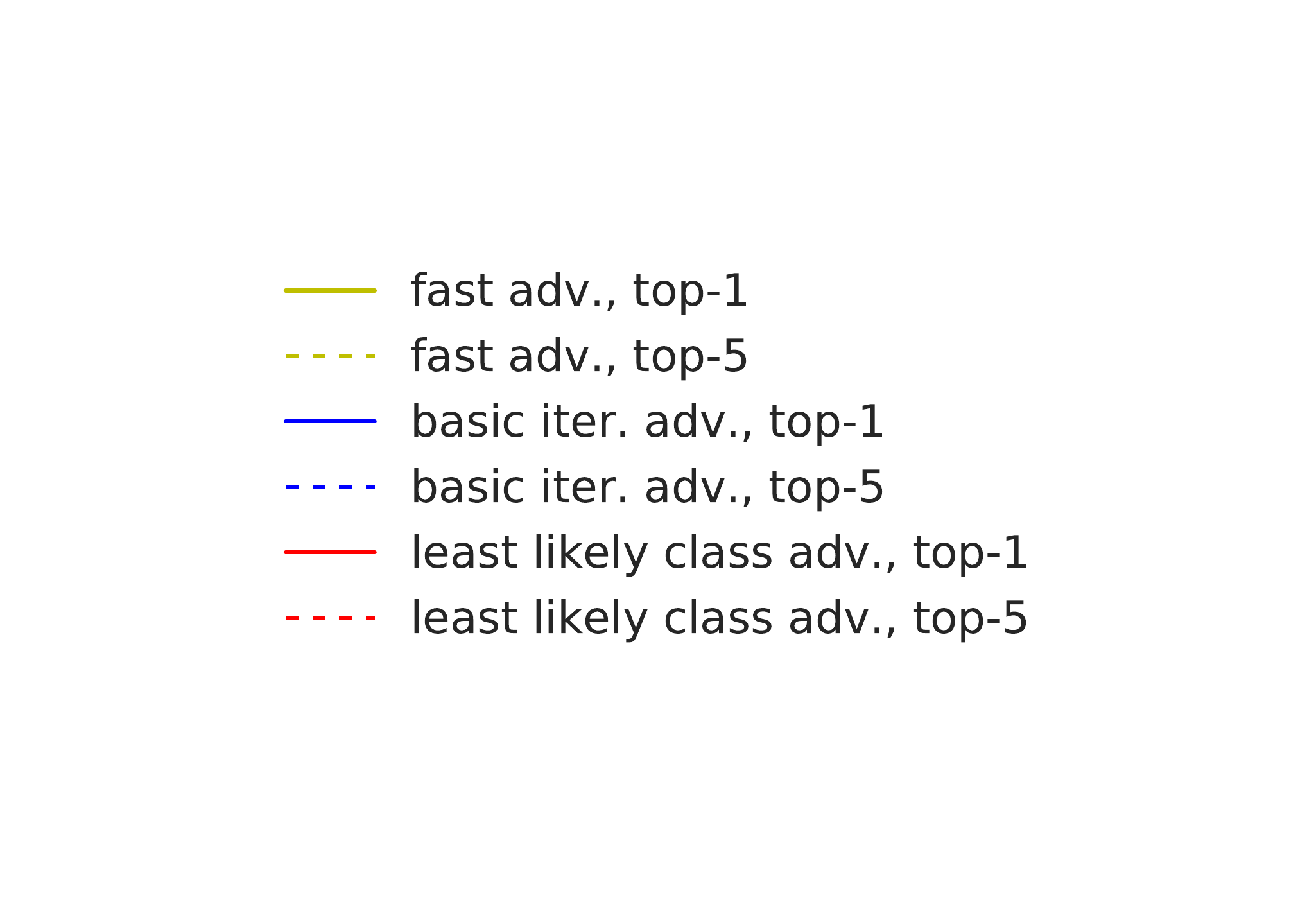}
  \end{subfigure}
  \caption{Comparison of adversarial destruction rates for various adversarial methods
  and types of transformations. All experiments were done with $\epsilon = 16$.
  }\label{fig-transform-destruction-rate}
\end{figure}  

\end{appendices}

\end{document}